\documentclass{article}
\pdfoutput=1
\if@natbib
  \RequirePackage{natbib}
\fi

\usepackage[final]{neurips_2024}
\usepackage{graphicx}
\usepackage{subfig}

\usepackage[utf8]{inputenc} 
\usepackage[T1]{fontenc}    
\usepackage{hyperref}       
\usepackage{url}            
\usepackage{booktabs}       
\usepackage{amsfonts}       
\usepackage{nicefrac}       
\usepackage{microtype}      
\usepackage{xcolor}         
\usepackage{amsmath}
\usepackage{amssymb}
\usepackage{booktabs}
\usepackage[capitalize]{cleveref}
\usepackage{xcolor}
\usepackage{bm}
\usepackage{bbm}
\usepackage{gensymb}
\usepackage{multirow}
\usepackage{makecell}

\title{Delta-LLaVA: Base-then-Specialize Alignment for Token-Efficient Vision-Language Models 

[Accepted at WACV 2026]}

\author{
Mohamad Zamini \! and \! Diksha Shukla, Senior Member, IEEE\\
University of Wyoming\\
{\tt\small mzamini@uwyo.edu, dshukla@uwyo.edu}
}

\begin{document}

\maketitle

\begin{abstract}
Multimodal Large Language Models (MLLMs) combine visual and textual representations to enable rich reasoning capabilities. However, the high computational cost of processing dense visual tokens remains a major bottleneck. A critical component in this pipeline is the visual projector, which bridges the vision encoder and the language model. Standard designs often employ a simple multi-layer perceptron for direct token mapping, but this approach scales poorly with high-resolution inputs, introducing significant redundancy. We present Delta-LLaVA, a token-efficient projector that employs a low-rank DeltaProjection to align multi-level vision features into a compact subspace before further interaction. On top of this base alignment, lightweight Transformer blocks act as specialization layers, capturing both global and local structure under constrained token budgets. Extensive experiments and ablations demonstrate that this base-then-specialize design yields consistent gains across multiple benchmarks with only $144$ tokens, highlighting the importance of token formation prior to scaling interaction capacity. With Delta-LLaVA, inference throughput improves by up to 
\textbf{55\%}, while end-to-end training 
accelerates by nearly \textbf{4--5$\times$} in pretraining
and over \textbf{1.5$\times$} in finetuning, 
highlighting the dual benefits of our design in both efficiency and scalability.
\end{abstract}

\section{Introduction}
\label{sec:intro}

Large language models have shown strong reasoning and generalization abilities. Recent multimodal models extends these capabilities to the visual domain. As universal interfaces, language models can follow natural language instructions while incorporating visual context. This enables general-purpose assistants that both perceive and reason about the world. Early multimodal systems such as Flamingo \cite{alayrac2022flamingo}, and more recent open-source efforts \cite{li2024enhancing, team2024gemma, touvron2023llama}, demonstrated that adding vision to pretrained language models greatly broadens the range of tasks they can solve. In most multimodal LLMs, images are converted into patch-based tokens by vision encoders like CLIP \cite{radford2021learning}. These tokens are projected into the language space and then processed alongside text tokens \cite{liu2023visual, liu2024improved}. While effective, this design often produces hundreds of tokens per image. As a result, visual processing becomes a major contributor to inference cost and latency.

A central bottleneck in this pipeline is the visual projector. Conventional projectors often adopt simple linear or MLP layers that retain all visual tokens, regardless of redundancy. This results in large token counts that overshadow textual prompts and inflate compute during both training and inference. Reducing the number of visual tokens has therefore become a primary research direction, with recent approaches proposing pooling layers, token merging, or lightweight attention mechanisms \cite{chen2024image, zhang2024cls, li2024inference}. However, these methods tend to compress features in ways that risk discarding fine-grained cues such as text regions, small objects, or structural layout. The trade-off between reducing tokens and maintaining semantic fidelity remains a major challenge.

\begin{figure*}[t]
  \centering
  \includegraphics[width=0.24\textwidth]{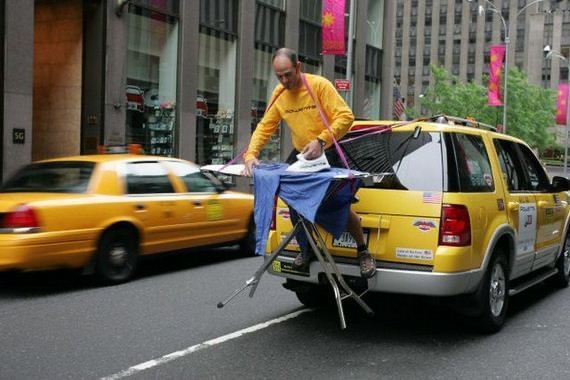}\hfill
  \includegraphics[width=0.24\textwidth]{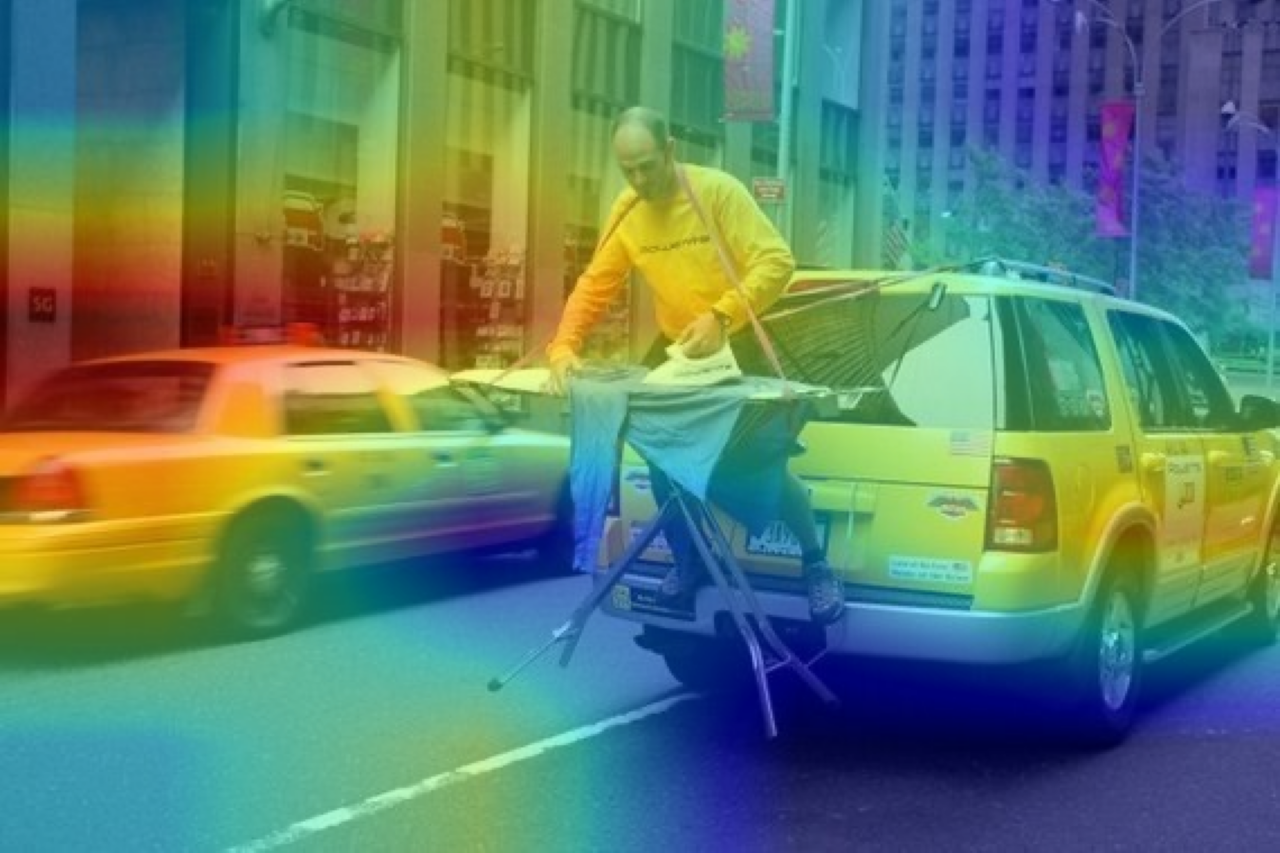}\hfill
  \includegraphics[width=0.24\textwidth]{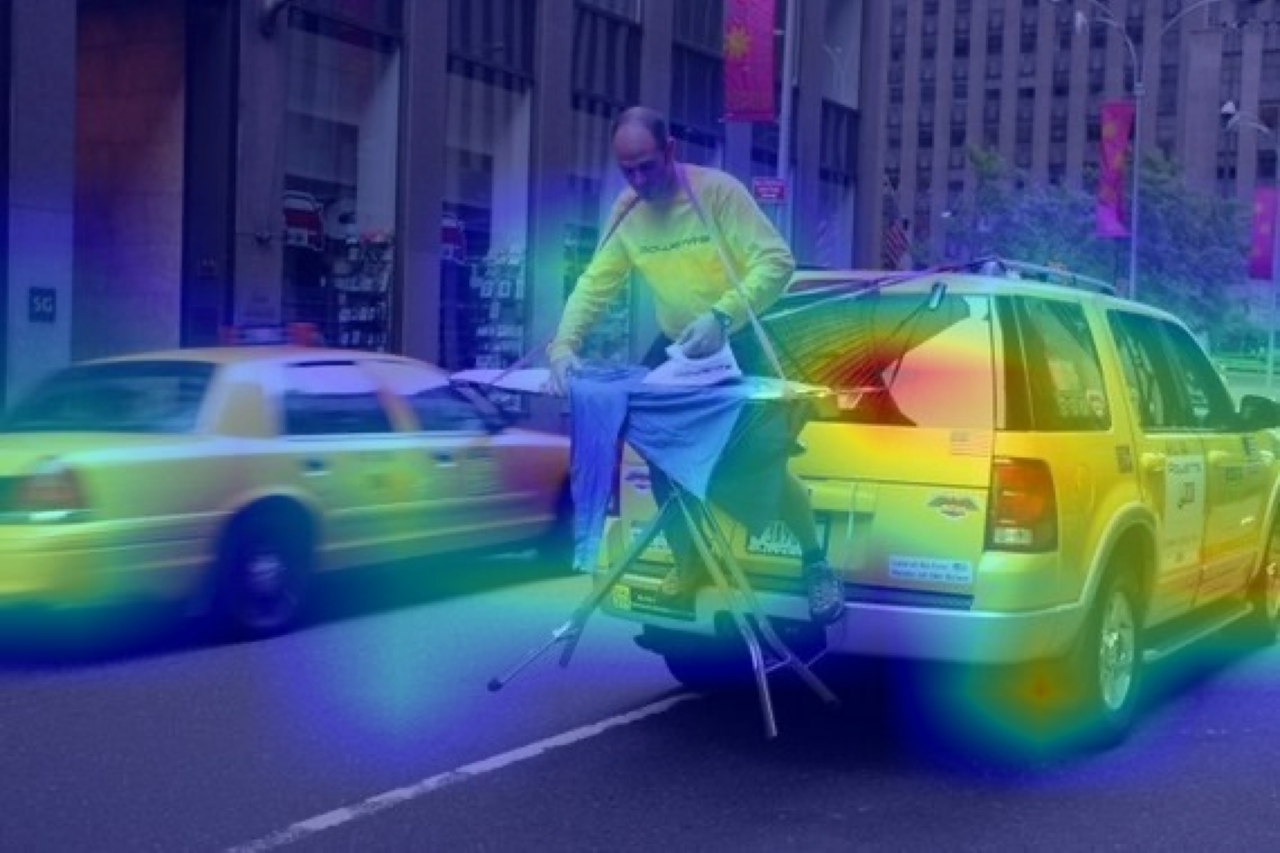}\hfill
  \includegraphics[width=0.24\textwidth]{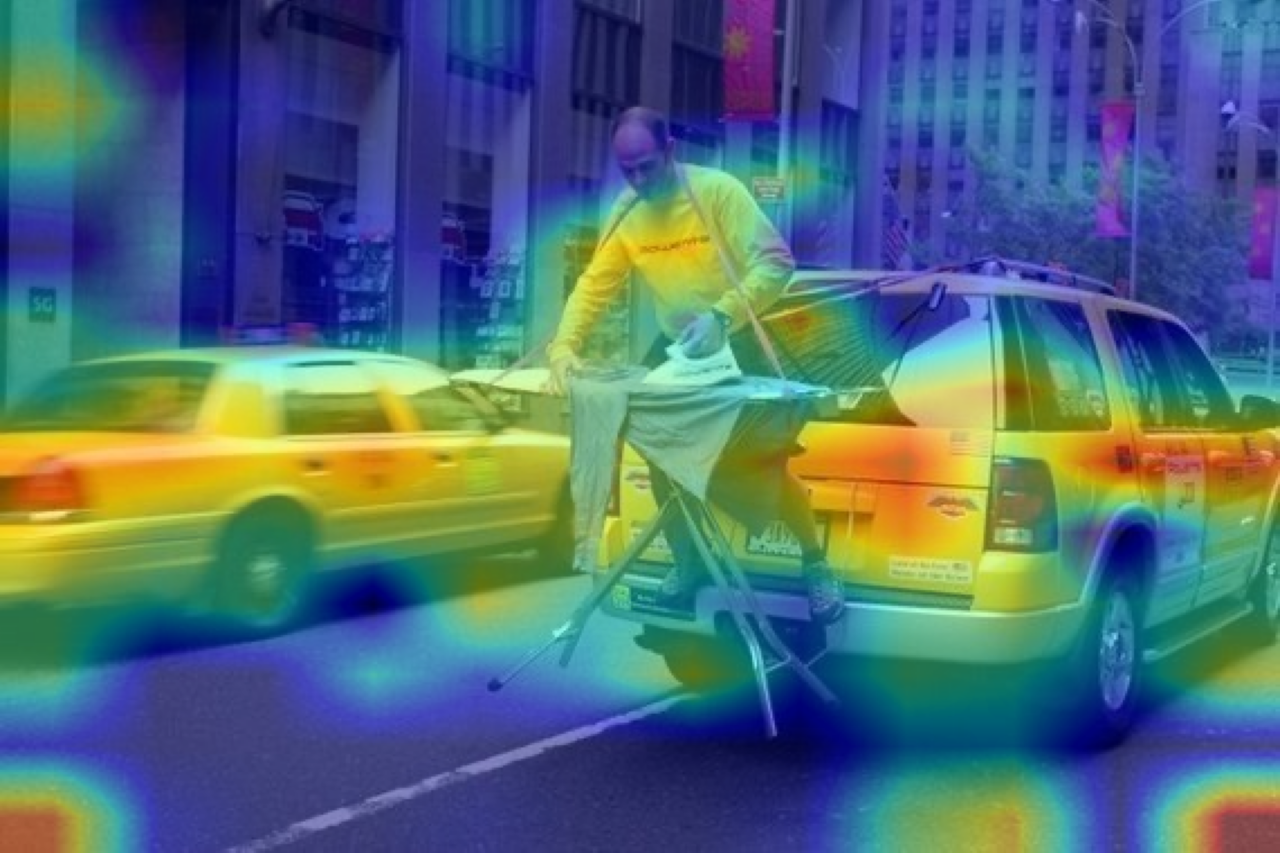}
  \caption{Grad-CAM visualization of the prompt: \emph{What is weird about this picture?} The response (144 tokens) is: \emph{The weird aspect of this picture is that a man is standing on the back of a yellow taxi while holding a clothes iron. This is unusual because it is not common to see someone using a clothes iron while standing on the back of a moving vehicle, especially a taxi. The man's actions seem unconventional and potentially dangerous, as he could lose his balance and fall off the taxi, causing injury or damage to the vehicle.} The images show results using 16, 64, and 144 visual tokens, from left to right (after the original image).}
  \label{fig:weird-picture}
\end{figure*}

In this paper, we introduce Delta-LLaVA, a multimodal projector that addresses this challenge through a base-then-specialize design. At its core is a low-rank DeltaProjection layer that aligns multi-level vision features into a compact token subspace before any further processing. This base alignment establishes the representational foundation on which subsequent modules operate. On top of DeltaProjection, we integrate two specialization layers: an efficient multi-head self-attention (EMHSA) block that captures long-range dependencies under compressed token budgets, and a transformer block (TB) that injects local structure and inductive bias. In this design, DeltaProjection provides the critical alignment that determines the quality of the aggregate reasoning, while EMHSA and TB serve to refine and specialize the aligned tokens for task-specific demands.

Our analyses show that DeltaProjection causes the largest aggregate performance improvements across reasoning benchmarks, while removing EMHSA or TB produces smaller, task-dependent changes. For example, EMHSA contributes to long-range reasoning robustness, and TB yields modest gains in structured question answering, but both are secondary to the role of DeltaProjection. This pattern highlights the importance of investing model capacity in alignment layers that determine how tokens are formed, before allocating additional compute to interaction layers that determine how tokens are used. By enforcing this ordering, \emph{Delta-LLaVA achieves competitive accuracy while substantially reducing FLOPS, advancing the efficiency of multimodal language models for real-world deployment.}
\section{Related Work}
\label{sec:related}
Recent multimodal LLMs build on pretrained language models by coupling them with vision encoders such as CLIP \cite{radford2021learning}, EVA-CLIP \cite{sun2023eva}, or SigLIP-2 \cite{tschannen2025siglip}. These systems typically rely on a projector to map visual features into the language embedding space, enabling joint reasoning over text and images \cite{liu2023visual, alayrac2022flamingo, li2023blip}. While effective, this design often produces hundreds of tokens per image, inflating the sequence length seen by the LLM. Sequence compression has long been studied in purely textual models. Funnel Transformers \cite{dai2020funnel} reduce sequence length through pooling, while pruning methods \cite{nawrot2022efficient} dynamically remove redundant tokens. However, redundancy is more pronounced in images, where adjacent patches often encode overlapping or low-information content. This makes visual token compression especially attractive for multimodal models.

\smallskip
\noindent
\textbf{Token pruning inside the LLM.}
Some works reduce inference cost by pruning visual tokens after they enter the LLM. LLaVolta \cite{chen2024llavolta} adaptively pools and drops tokens across transformer layers, while EViT \cite{liang2022not} and Dynamic-ViT \cite{rao2021dynamicvit} remove tokens within the vision backbone or multimodal stack. These methods save computation but require modifying or retraining the backbone, reducing modularity.

\smallskip
\noindent
\textbf{Hierarchical and adaptive compression.}
Some other studies propose adaptive or hierarchical compression. MQT \cite{hu2024matryoshka} and M3 \cite{cai2024matryoshka} employ Matryoshka representation learning \cite{kusupati2022matryoshka} to generate nested representations that adjust compute budgets at inference time. While flexible, these approaches may still propagate redundant tokens when operating at higher budgets.

\smallskip
\noindent
\textbf{Projector-side compression.}
A parallel direction which is our point of interest, compresses tokens before they reach the LLM. LLaVA-PruMerge \cite{shang2024llava} selects salient patches using attention and merges redundant ones by clustering. Honeybee \cite{cha2024honeybee} introduces a compact bottleneck, while TokenPacker \cite{li2024tokenpacker} uses coarse-to-fine attention to balance global and local detail. These methods are modular and LLM-agnostic, but still face a trade-off between efficiency and preservation of fine-grained semantics such as small text or layout.

Existing methods primarily frame compression as pruning or merging, with the risk of discarding task-critical information. In contrast, our work treats the projector as an alignment module rather than just a bottleneck. In our model, DeltaProjection as a base alignment layer, establishes a compact and semantically faithful token subspace. Ablations confirm that DeltaProjection is the principal driver of aggregate reasoning performance, while EMHSA and TB yield complementary, task-specific gains. This alignment-first, specialization-second perspective distinguishes our approach from prior projector-side compression methods.

\section{Method}
Multimodal LLMs aim to generate instruction-following responses from visual and textual inputs. They are generally composed of three key components:

\begin{itemize}
\item \textbf{Visual Encoder} $\mathcal{F}_I$: This module encodes an input image $\mathbf{I}_{\text{img}} \in \mathbb{R}^{H \times W \times 3}$ into a sequence of visual embeddings $\mathbf{I}_v \in \mathbb{R}^{N \times C}$. Most current MLLMs employ CLIP-ViT-L/14 as the backbone, with patch size $P=14$, resulting in $N = HW/P^2$ visual tokens.

\item \textbf{Visual Projector} $\Gamma_{I \rightarrow T}$: The visual projector serves as a crucial bridge between the vision and language models by translating visual features into visual tokens within a text embedding space that the language model can interpret. It transforms visual embeddings $\mathbf{I}_v$ into a set of visual tokens $\mathbf{T}_v$ within the language model’s embedding space $\mathbb{T}$, aligning the modalities for joint processing.

\item \textbf{Language Model} $\Phi(\mathbf{T}_v, \mathbf{T}_t)$: The LLM receives both visual tokens $\mathbf{T}_v$ and textual tokens $\mathbf{T}_t$, and generates responses autoregressively. Given a target sequence $Y = \{y_i\}_{i=1}^L$, the generation probability is defined as:
\begin{equation}
    p(Y \mid \mathbf{T}_v, \mathbf{T}_t) = \prod_{i=1}^L p(y_i \mid \mathbf{T}_v, \mathbf{T}_t, y_{<i})
\end{equation}
\begin{figure}[t]
  \centering
  \includegraphics[width=0.5\textwidth]{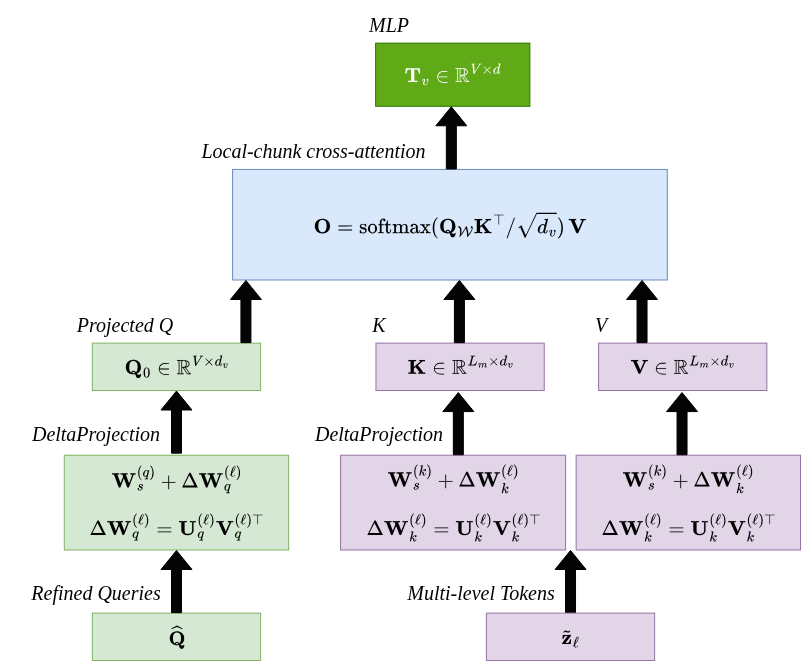}
  \caption{DeltaProjection inside Delta-LLaVA.}
\end{figure}
\end{itemize}
While standard MLLM architectures are effective, their inference cost is jointly dictated by the number of visual tokens \(N\) and the size of the language model \(\Phi\). Importantly, the computational cost of \(\Phi(T_v, T_t)\) scales quadratically with the total number of tokens, making token count a critical factor in runtime efficiency. In this setting, the visual projector plays a central role: it transforms the \(N\) visual embeddings \(I_v\) into a set of \(M\) visual tokens \(T_v\), which are passed to the language model.

Minimizing the number of visual tokens (\(M < N\)) is therefore an essential strategy for improving inference efficiency. To this end, we introduce Delta-LLaVA, a visual projector that bridges the vision encoder and language model using as few tokens as possible. 

\section{Delta-LLaVA: Multimodal Projector}
To reduce the number of visual tokens while preserving semantic fidelity, we introduce a Base-then-Specialize visual projection module that couples lightweight convolutional blocks with transformer-style attention. The Delta projector receives two streams from the vision tower: dense patch embeddings $\mathbf I_v=\{\mathbf z_p\}_{p=1}^{N}$ and a compact multi-level summary $\mathbf I_m=\{\tilde{\mathbf z}_\ell\}_{\ell=1}^{L_m}$.  Its task is to produce a {\it compressed} representation $\mathbf Y\in\mathbb R^{V\times d}$, with $V\!\ll\!N$, retaining the cues most predictive for the language model $\Phi$.

\smallskip
\noindent
\subsection{Query construction.}%
\quad
Let $\mathbf Z$ denote patch-wise features on a raw $\frac{H}{P}\times\frac{W}{P}$ grid. 
We first downsample this grid to $\frac{H}{Ps}\times\frac{W}{Ps}$ via an interpolation operator 
$Interp(\cdot)$, where $s \geq 1$ is the spatial scale. 
This preserves positional correspondences while reducing the query grid size. At the core of Delta-LLaVA is DeltaProjection, a low-rank projection inspired by a parameter-efficient adaptation method \cite{mikaelyan2025deltallm}. 

Once alignment is established, Delta-LLaVA applies two specialization blocks for refining queries: 
1) Multi-Head Convolutional Attention and 2) Efficient Multi-Head Self-Attention for capturing long-range dependencies with optional spatial reduction, enabling global context reasoning under compressed token budgets. 
Finally, we add fixed 2D sinusoidal positional embeddings.

\smallskip
\noindent
\textbf{Multi-Head Convolutional Attention (MHCA).}
In our implementation, MHCA is a lightweight convolutional mixing block that
approximates local attention while avoiding explicit dot–product computations.
Instead of computing queries, keys, and values followed by a softmax, the input
is processed by grouped $3\times 3$ depthwise convolutions (with the number of
groups matching the number of heads), so that each head captures local spatial
context in its own channel subset. The result is normalized with LayerNorm,
passed through a ReLU activation, and then merged across heads with a convolution. A residual connection adds the input back to the output.

This design preserves the multi–head flavor of attention—each head aggregates a
different local receptive field—while keeping the complexity linear in $HW$.
MHCA therefore provides efficient locality modeling aligned with CNN inductive
biases, without the quadratic overhead of token–token attention.

\begin{figure*}[t]
  \centering
  \subfloat[Inference efficiency of the models on 100 samples. Projector FLOPs are $\leq 0.02$.]{
    \includegraphics[width=0.48\textwidth]{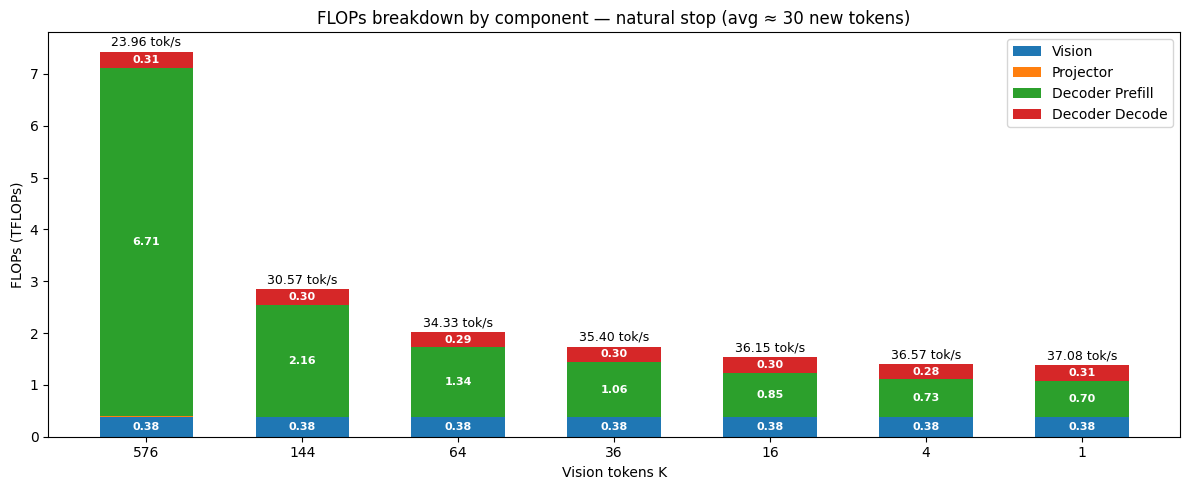}
    \label{fig:prune-flops}
  }
  \hfill
  \subfloat[Delta-LLaVA 144 tokens surpasses LLaVA--1.5 on GQA.]{
    \includegraphics[width=0.48\textwidth]{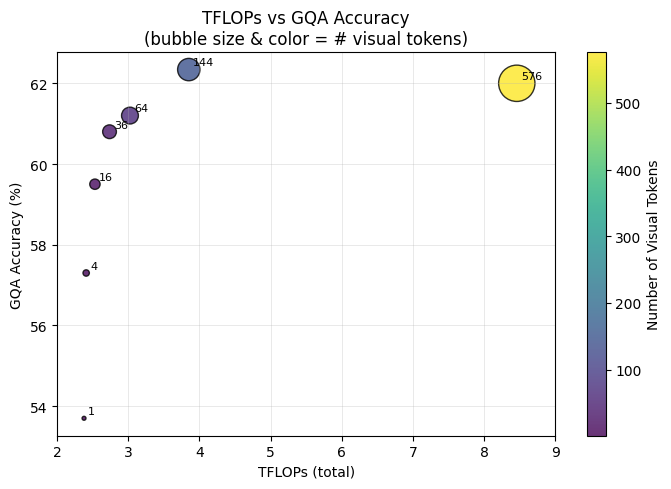}
    \label{fig:gqa}
  }
  \caption{Inference performance and accuracy with naturally terminated generations ($\approx$30 new tokens on average) under varying visual-token counts.}
  \label{fig:flops-gqa}
\end{figure*}

\smallskip
\noindent
\textbf{Efficient Multi-Head Self-Attention (EMHSA).}
To capture global semantics, we adopt a stabilized multi-head self-attention with
pre-normalization and optional spatial reduction. Given normalized input 
$\mathbf X \in \mathbb R^{B\times N\times C}$, we form
$\mathbf Q,\mathbf K,\mathbf V \in \mathbb R^{B\times N\times C}$ and split them
into $H$ heads of dimension $d=C/H$. The attention is computed as
\begin{equation}
\mathrm{Attn}(\mathbf Q,\mathbf K,\mathbf V)
=softmax\!\Bigl(\frac{\mathbf Q \mathbf K^\top}{\sqrt{d}}\Bigr)\mathbf V.
\end{equation}

For efficiency, when $s>1$ we downsample $\mathbf K,\mathbf V$ to
$\mathbf K',\mathbf V' \in \mathbb R^{B\times (N/s^2)\times C}$, reducing
complexity from $O(HN^2)$ to $O(HN\cdot N/s^2)$ while preserving global
receptive fields.\vspace{-0.8em}

\smallskip
\noindent
\subsection{DeltaProjection.}
The refined queries $\widehat{\mathbf Q}$ are then mapped into the vision--language
embedding space with a DeltaLLM-style low-rank projection:
\begin{align}
\widehat{\mathbf Q}
&= \operatorname{Refine}\!\Bigl(\operatorname{Interp}(\mathbf Z,\tfrac{H}{Ps},\tfrac{W}{Ps})\Bigr) + \text{Pos2D}, \\[3pt]
\mathbf Q_0
&= \bigl(\mathbf W_s + \Delta \mathbf W_q^{(\ell)}\bigr)\,\widehat{\mathbf Q}, 
\qquad
\Delta \mathbf W_q^{(\ell)} = \mathbf U_q^{(\ell)} \mathbf V_q^{(\ell)\top}.
\end{align}
Here, $\mathbf W_s\in\mathbb R^{d_v\times C}$ is a shared base weight, and
$\Delta \mathbf W_q^{(\ell)}$ is a lightweight, rank-$r$ update specific to
layer $\ell$. The number of queries is 
\begin{equation}
    V = \tfrac{HW}{P^2 s^2}.
\end{equation}

\subsection{Coarse-to-fine mixing.}
The projected queries $\mathbf Q_0$ are then enriched by a cascade of transformer
blocks (EMHSA $\to$ MHCA $\to$ MLP), followed by low-rank projection and local
chunk attention:
\begin{equation}
\mathbf Q = \operatorname{NTB}(\mathbf Q_0).
\end{equation}
This cascade preserves resolution $V$ while infusing each token with both local
detail and long-range semantics.

This cascade preserves resolution $V$ but enriches each token with information
from both its immediate neighborhood and the full patch grid, enabling reasoning
about fine details (object attributes, small text) while still capturing
scene-level context.

\smallskip
\noindent
\textbf{Key–value pathway.}%
\quad
In parallel to query construction, a compact set of multi-level feature tokens
$\tilde{\mathbf z}_\ell$ is transformed into keys and values via the same
low-rank $\Delta$-projection used for queries. Let $M=L_m$ denote the number of
memory tokens. We obtain
\begin{align}
\mathbf K &= \bigl[\mathbf k_\ell\bigr]_{\ell=1}^{M}, &
\mathbf V &= \bigl[\mathbf v_\ell\bigr]_{\ell=1}^{M}, \\
\mathbf k_\ell &= \operatorname{Proj}_k(\tilde{\mathbf z}_\ell), &
\mathbf v_\ell &= \operatorname{Proj}_v(\tilde{\mathbf z}_\ell),
\end{align}
where each $\operatorname{Proj}_\cdot$ is a shared base matrix plus a rank-$r$
Delta-LLaVA update (DeltaLLM-style \cite{mikaelyan2025deltallm}), yielding parameter- and compute-efficient
adaptation.

\smallskip
\noindent
\textbf{Windowed cross-attention.}%
\quad
inspired by TokenPacker \cite{li2024tokenpacker}, let the refined query sequence have length $V=g^2$ (with $g$ the query grid
size after interpolation). We partition the query grid into non-overlapping
windows $\{\mathcal W_i\}$ of size $w\times w$, so there are $V/w^2$ windows.
Consistent with the implementation, keys and values are downsampled/partitioned
by the same spatial scale $s$ to yield
window-aligned subsets $\mathbf K_{\mathcal W_i},\mathbf V_{\mathcal W_i}$.
Within each window we compute
\begin{align}
\mathbf O_{\mathcal W_i}
&=
\operatorname{softmax}\!\left(
  \tfrac{\mathbf Q_{\mathcal W_i}\mathbf K_{\mathcal W_i}^\top}{\sqrt d}
\right)\mathbf V_{\mathcal W_i},
\label{eq:cross_attn_local}
\\[3pt]
\mathbf Y
&=
\bigl[\mathbf O_{\mathcal W_i}\bigr]_{i=1}^{V/w^{2}},
\end{align}
where $d$ is the per-head dimension. This \emph{local chunk attention} reduces
quadratic cost by restricting interactions to spatial neighborhoods while
anchoring every window to the compact multi-level memory.
\begin{table*}[ht]
\centering
\caption{Comparison of token compression methods across varying compression rates. All models use the Vicuna-1.5 7B backbone unless otherwise specified. The best and second-best results are highlighted in bold and \underline{underline}, respectively. Our method achieves superior performance under extreme compression and remains competitive at moderate levels. Results for 1–36 tokens are taken from \cite{li2024inference}; the rest are reported from the original papers. A dash (–) indicates that results were not reported for the corresponding benchmark.} 
\small
\setlength{\tabcolsep}{4.2pt}
\begin{tabular}{lccccccccc}
\toprule
\textbf{Method} & \textbf{\# Token} & \textbf{GQA} & \textbf{MMB} & \textbf{MME} & \textbf{POPE} & \textbf{SQA} & \textbf{TextVQA} & \textbf{VizWiz} & \textbf{VQAv2} \\
\midrule
LLaVA-1.5 & 576 & \underline{62.0} & 64.3 & \textbf{1510.7} & 85.9 & 66.8 & \textbf{58.2} & 50.0 & \textbf{78.5} \\
TokenPacker  & 144 & 61.9 & 65.1 & -- & \textbf{87.0} & -- & -- & 52.0 & \underline{77.9} \\
Matryoshka Multi. & 144 & 61.3 & \textbf{66.4} & -- & \textbf{87.0} & -- & -- & \underline{53.1} & -- \\
Matryoshka Query & 144 & 61.4 & 64.4 & 1446.5 & 83.9 & \underline{67.5} & -- & 52.0 & 76.4 \\
\textbf{Delta-LLaVA (Ours) } & 144 & \textbf{62.34 }& \underline{65.95} & \underline{1466.52} & \underline{86.77} & \textbf{68.7} & 57.14 & \textbf{56.6} & 76.1 \\
\midrule
\textbf{LLaVA-1.5 13B} & 576 & \textbf{63.3} & \textbf{67.7} & \textbf{1531.3} & 86.2 & \textbf{71.6} & \textbf{61.3} & 53.6 & \textbf{80.0} \\
\textbf{Delta-LLaVA-13B (Ours) } & 144 & 62.7 & 67.4 & 1527.41 & \textbf{87.3} & \textbf{71.6} & 59.25 & \textbf{58.2} & 76.9 \\
\midrule
PruMerge & $\sim$32 & 57.2 & 60.9 & 1350.3 & 76.3 & 68.5 & \textbf{56.0} & 45.2 & 72.0 \\
TokenPacker & 36 & 59.6 & 62.8 & \underline{1440.9} & 83.3 & \textbf{71.0} & 53.2 & 50.2 & \underline{75.0} \\
Matryoshka Multi. & 36 & 60.3 & \textbf{64.8} & -- & \underline{85.5} & -- & -- & \underline{52.8} & -- \\
Matryoshka Query & 36 & 58.8 & 63.4 & 1416.3 & 81.9 & 66.8 & -- & 51.0 & 73.7 \\
QueCC & 36 & \underline{60.5} & 62.5 & \textbf{1442.0} & 84.5 & \underline{70.6} & 53.3 & 50.1 & \textbf{75.8} \\
\textbf{Delta-LLaVA (Ours)} & 36 & \textbf{61} & \underline{64.4} & 1424.7 & \textbf{85.7} & 68.6 & \underline{54.4} & \textbf{56.2} & 74.3 \\
\midrule
TokenPacker & 16 & 58.9 & \underline{62.7} & 1378.8& \underline{83.7} & 68.1 & \underline{52.5} & 50.5 & \underline{74.4} \\
Matryoshka Query & 16 & 57.6 & 61.9 & \textbf{1408.5} & 80.8 & 67.5 & -- & 49.8 & 71.1 \\
QueCC & 16 & \underline{59.0} & 62.2 & \underline{1408.0} & 83.4 & \textbf{70.7} & 51.3 & 47.7 & \textbf{74.5} \\
\textbf{Delta-LLaVA (Ours) } & 16 & \textbf{59.5 }& \textbf{62.9} & 1375.9 & \textbf{84.7} & \underline{69.7} & \textbf{53.6} & \textbf{55.2} & 73.1 \\
\midrule
TokenPacker & 4 & 56.2 & 61.5 & \underline{1347.6} & 81.7 & \underline{68.5}& \underline{49.2} & 45.7 & 70.5 \\
Matryoshka Query & 4 & 53.0 & 56.5 & 1176.1 & 77.6 & 65.1 & -- & \underline{49.4} & 64.1 \\
QueCC & 4 & \underline{56.5} & \textbf{62.1} & \textbf{1390.3} & \underline{81.8} & \textbf{68.6} & 48.7 & 45.0 & \underline{70.6} \\
\textbf{Delta-LLaVA (Ours)} & 4 & \textbf{57.3} & \underline{61.7} & 1314.2 & \textbf{82.2} & 67.9 & \textbf{53.1} & \textbf{52.6} & \textbf{72.0} \\
\midrule
TokenPacker & 1 & 53.4 & 58.7 & \textbf{1262.4} & 80.7 & \underline{69.4} & 46.2 & 41.1 & \textbf{66.9} \\
Matryoshka Multi. & 1 & 52.6 & \textbf{59.5} & -- & 78.4 & -- & -- & \underline{49.4} & -- \\
Matryoshka Query & 2 & 50.8 & 54.4 & 1144.0 & 74.5 & 65.0 & -- & 48.5 & 61.0 \\
QueCC & 1 & \underline{53.5} & \underline{59.4} & \underline{1269.1} & \underline{81.3} & \textbf{69.9} & \underline{46.8} & 44.1 & \underline{67.3} \\
\textbf{Delta-LLaVA (Ours)} & 1 & \textbf{53.7} & 57.6 & 1257.9 & \textbf{82.4} & 68.7 & \textbf{48} & \textbf{51} & 66.1 \\
\bottomrule
\end{tabular}

\label{tab:main_results}
\end{table*}
\begin{itemize}
  \item \textbf{Low-rank projections.} Both $\operatorname{Proj}_k$ and
  $\operatorname{Proj}_v$ share the same $\Delta$-form as the query projection
  (shared base + rank-$r$ update).
  \item \textbf{Compact memory.} Since $M \ll V$, the key–value cache remains
  small relative to the query sequence, keeping the memory branch lightweight.
  \item \textbf{Efficiency.} Combining localized attention with a compact
  memory provides fine spatial reasoning and long-range context at much lower
  cost than global self-attention.
\end{itemize}

\smallskip
\noindent
\subsection{Feed-forward refinement.} Finally, a position-wise two-layer MLP with hidden width $h=4096$ refines the
tokens:
\begin{equation}
\mathbf T_v
\;=\;
\mathbf T_v \;+\;
\mathbf W_2\,\sigma\!\bigl(\mathbf W_1\,\operatorname{LN}(\mathbf T_v)\bigr).
\label{eq:ffn}
\end{equation}
This residual feed-forward path enhances non-linear expressivity while
maintaining stability.

\smallskip
\noindent
\section{FLOPs Analysis.}%
\quad
Let $V_0=\tfrac{HW}{P^2}$ be the raw patch-token count and $V=\tfrac{V_0}{s^2}$ the
post-compression tokens at spatial scale $s$. We separate vision/projector from LLM prefill/decode.

\vspace{-0.8em}\paragraph{Projector.}
The proposed Delta-LLaVA projector includes low-rank projections,
convolutional blocks, windowed attention, and feed-forward refinement. 
For fixed embedding dimension $d$, rank $r$, and local window size $L_m$, the 
dominant terms are
\begin{itemize}
  \item low-rank projections: $\mathcal{O}(V d)$,
  \item convolutional mixing (MHCA): $\mathcal{O}(V d k^2)$,
  \item feed-forward MLP: $\mathcal{O}(V r d^2)$,
  \item local/windowed attention: $\mathcal{O}(V L_m d)$.
\end{itemize}
Each term scales \emph{linearly} with $V$, so the overall projector complexity
is
\begin{equation}
    F_{\text{proj}}(s) = \Theta\!\left(V\right) = \Theta\!\Bigl(\tfrac{V_0}{s^2}\Bigr).
\end{equation}

For
fixed $d$ (embed dim), expansion ratio $r$, kernel size $k$, and local window
$L_m$, the dominant terms are

\begin{equation}
    \mathcal{O}(V d) \;+\; \mathcal{O}(V d k^2) \;+\; \mathcal{O}(V r d^2) \;+\; \mathcal{O}(V L_m d)
\;=\; \Theta(V).
\end{equation}

Thus, doubling the scale factor $s$ reduces projector FLOPs by a factor of
four, in contrast to global self-attention which scales quadratically
$\Theta(V^2 d)$.

\paragraph{LLM Prefill.}
With $T$ text tokens, the prefill context length is $S_0=T+V$. Per layer,
\begin{equation}
    F_{\text{prefill}}(T,V)\;=\;\Theta\!\big(S_0\, r d^2\big)\;+\;\Theta\!\big(S_0^2 d\big).
\end{equation}
In 7B-class models the MLP term $\Theta(S_0 r d^2)$ dominates,
so prefill FLOPs scale approximately linearly with $V$.
This matches our measurements: $F_{\text{prefill}}$ drops from $6.72$ TFLOPs
at $V{=}576$ to $2.16$ (@$144$), $0.85$ (@$16$), and $0.70$ (@$1$).

\paragraph{LLM Decode.}
With KV caching, generating $G$ tokens yields per-layer cost
\begin{equation}
    F_{\text{decode}}(T,V,G)\;=\;\Theta\!\big(G\, r d^2\big)\;+\;\Theta\!\big(G\, S_0 d\big).
\end{equation}

Because the MLP term dominates, $F_{\text{decode}}$ is nearly constant in $V$.
In our natural-stop runs ($G{\approx}30$ on average), decode FLOPs range only
from $0.31$ TFLOPs (@$576$) to $0.28$ (@$4$–$16$). 
Runtime per generated token shows modest variation: tokens/sec rises from
$\approx 24$ (@$576$) to $\approx 37$ (@$1$), a $\sim55\%$ gain,
even though decode FLOPs remain flat.

\vspace{-0.8em}\paragraph{Vision Encoder.}
$F_{\text{vision}}$ is effectively constant (e.g., $0.382$ TFLOPs in all settings),
since compression happens post-backbone.

The total complexity is
\begin{equation}
\begin{aligned}
F_{\text{total}}(V) \approx\; & F_{\text{vision}} + F_{\text{proj}}(V) + F_{\text{prefill}}(T,V) \\
& {}+ F_{\text{decode}}(T,V,G).
\end{aligned}
\end{equation}

Reducing $V$ meaningfully cuts projector and \textit{prefill} FLOPs (and KV memory),
but \textit{decode} FLOPs stay nearly constant and decode \emph{runtime} is limited by
KV-cache read bandwidth. For long generations ($G \gg T+V$), decode dominates inference time,
so reducing $V$ has negligible impact on end-to-end latency.

\section{Experiments}
\textbf{Model Setup.} We adopt LLaVA-1.5 as the foundation for our multimodal language models (MLLMs), integrating various language backbones including Vicuna-7B, Vicuna-13B, Qwen-7B. For vision encoding, we employ CLIP-ViT-B/32 (224) and CLIP-ViT-L/14 (336). All models are trained for one epoch using 8 $\times$ NVIDIA H100 GPUs.

\smallskip
\noindent
\textbf{Evaluation Benchmarks.} We follow a two-stage training pipeline. The Delta-LLaVA module is first trained on CC-558K \cite{liu2023visual} for modality alignment. We then apply instruction tuning using the 665K LLaVA mixture \cite{liu2024improved}. Evaluation is performed on a diverse suite of benchmarks using LMMs-Eval tool \cite{zhang2025lmms}: general VQA tasks ($VQA^{v2}$ \cite{goyal2017making}, GQA \cite{hudson2019gqa}, VizWiz \cite{gurari2018vizwiz}); hallucination detection (POPE \cite{li2023evaluating}); comprehensive reasoning benchmarks (MMBench \cite{liu2024mmbench}, MME \cite{liang2024survey}); and image captioning (Nocaps \cite{agrawal2019nocaps} and Flicker30K\cite{plummer2015flickr30k}).

\subsection{Empirical Analysis.}
Table \ref{tab:main_results} compares token compression strategies across a wide range of visual token counts. At the full setting of 576 tokens, the baseline LLaVA-1.5 achieves strong performance, particularly on MME (1510.7) and VQAv2 (78.5). However, as tokens are reduced, most baselines experience steep accuracy drops, especially below 16 tokens. Our method, Delta-LLaVA, consistently outperforms or closely matches competing approaches across all compression levels. At moderate compression (144 tokens), Delta-LLaVA improves GQA (62.34 vs. 61.9 TokenPacker) and SQA (68.7 vs. 67.5 Matryoshka Query), while preserving competitive scores on MME and POPE. At aggressive compression (36 and 16 tokens), Delta-LLaVA delivers the best or second-best results in the majority of benchmarks, notably maintaining VizWiz performance (56.2 at 36 tokens, 55.2 at 16 tokens) where others degrade. Even at extreme settings of 4 or 1 token, Delta-LLaVA remains robust, achieving the top scores on GQA, POPE, TextVQA, and VizWiz, highlighting its resilience under severe token bottlenecks. Notably, scaling Delta-LLaVA to a 13B backbone further boosts overall accuracy, suggesting complementary gains from larger language capacity. These results demonstrate that Delta-LLaVA achieves state-of-the-art trade-offs between efficiency and performance, particularly excelling at extreme compression where prior methods collapse.

\begin{table}[h]\vspace{-0.4em}
\centering
\footnotesize
\begin{tabular}{lcccccccc}
\toprule
\textbf{Method} & \textbf{\# Token} & \textbf{GQA} & \textbf{MMB} & \textbf{MME} & \textbf{POPE} & \textbf{SQA}  & \textbf{VizWiz}  \\
\midrule
Qwen-VL-Chat & 448 & \underline{57.5} &  60.6 & \underline{1487.5} & 68.2   & \underline{38.9}  & \textbf{78.2} \\
\textbf{Delta-LLaVA Qwen} & 144 & \textbf{61.77} & \textbf{72.0} & \textbf{1524.0} &\textbf{ 87.0} & \textbf{73.1} & \textbf{41.7}  \\
\textbf{Delta-LLaVA Qwen} & 36 & 55.1 & \underline{67.3} & 1452.3 &\underline{86.2} & \underline{72.4} & 33.1 \\
\midrule
\end{tabular}
\caption{Results obtained using the LLaVA architecture with Qwen-7B as the language model compared to Qweun-VL-Chat.}
\label{tab:qwen}
\end{table}
Table \ref{tab:qwen} compares Delta-LLaVA with Qwen-7B as the language model against the baseline Qwen-VL-Chat. Despite using significantly fewer visual tokens (144 vs. 448), Delta-LLaVA Qwen delivers clear improvements across most benchmarks, achieving the best scores on GQA, MMB, MME, POPE, SQA, and VizWiz. Notably, the 144-token variant surpasses the baseline by more than 3 points on GQA and over 11 points on MMB, while also increasing MME to 1524.0. Even under extreme compression to 36 tokens, Delta-LLaVA Qwen maintains competitive performance, with only modest drops in accuracy relative to the full-token baseline. These results demonstrate that the Delta-LLaVA framework can drastically reduce the visual token budget while enhancing reasoning performance, highlighting both efficiency and robustness compared to Qwen-VL-Chat.
\begin{table}[h]
\centering
\begin{tabular}{lcccccc}
\toprule
\textbf{Model} & \textbf{GQA} & \textbf{MMB} & \textbf{MME} &  \textbf{POPE} & \textbf{SQA} \\
\midrule
No EMHSA           & 61.07 & \underline{64.77} & \underline{1449.84}  & \textbf{88.0} & 68.17 \\
No DeltaProj & 61.47 & 62.71 & 1431.67  & 85.57 & 68.32 \\
No TB             & \underline{62.0} & 64.17 & 1422.42  & \underline{86.83} & \textbf{70.20} \\
Full Model             & \textbf{62.34} & \textbf{65.95} & \textbf{1466.52} & 86.77 & \underline{68.67} \\
\bottomrule
\end{tabular}
\caption{Ablation study across reasoning and hallucination benchmarks. We removed one module and kept the rest each time. Bold values indicate the best performance within each column.}
\label{tab:ablation}\vspace{-0.8em}
\end{table}
\subsection{Ablation Study}
To evaluate the contribution of each component in our proposed architecture, we conducted an ablation study by selectively removing individual modules. Table \ref{tab:ablation} reports performance across five benchmarks: GQA, MMB (Reasoning), MME (Reasoning), POPE (Hallucination), and SQA. 

The results highlight that each module contributes differently. Removing E-MHSA reduces reasoning performance (MMB, MME) but yields the highest POPE score, suggesting that global self-attention aids reasoning but may also amplify hallucinations. Eliminating DeltaProjection consistently lowers reasoning metrics, demonstrating the importance of low-rank query alignment for efficient cross-modal fusion. Removing TB slightly decreases MME performance but surprisingly improves SQA, indicating that some linguistic tasks benefit from a simplified local representation. Overall, the full model achieves the best balance across tasks, with strong reasoning ability (MMB, MME) and competitive hallucination control (POPE).

\subsection{Training Efficiency}
\label{sec:training-speed}
We report end-to-end wall-clock time for both pretraining and instruction
finetuning across different Delta-LLaVA variants in Table \ref{tab:training-speed}.
All experiments were conducted under identical hardware and dataloader
configurations to isolate the effect of token compression on efficiency.
\begin{table}[h]
  \centering
  \begin{tabular}{lcc}
    \toprule
    \textbf{Model} & \textbf{Pretraining} & \textbf{Finetuning} \\
    \midrule
    LLaVA 576 & 1:24:01 & 4:34:20 \\
    LLaVA 144 & 0:33:32 & 3:21:37 \\
    LLaVA 64 & 0:23:24 & 3:17:23 \\
    LLaVA 16 & 0:18:36 & 3:14:29 \\
    LLaVA 4 & 0:16:32 & 3:12:34 \\
    LLaVA 1 & 0:16:00 & 2:51:50 \\
    \bottomrule
  \end{tabular}
    \caption{Training speed comparison across variants. 
  Reported as wall-clock time for one epoch of pretraining and finetuning.}
  \label{tab:training-speed}
  \vspace{-0.8em}
\end{table}

We observe a clear monotonic reduction in training time as the number of
visual tokens decreases. Standard LLaVA requires more than 3.5 hours
for pretraining and over 7.5 hours for finetuning per epoch, whereas
our most compressed variant (LLaVA~1) reduces these to 16 minutes and
2.9 hours, respectively. This corresponds to a $\sim\!13\times$
speedup in pretraining and more than $2.5\times$ acceleration in finetuning.

Interestingly, the majority of gains are realized during the pretraining stage,
where visual encoding dominates compute. As token count is reduced (e.g.,
from 144 down to 1), the quadratic attention cost shrinks substantially,
leading to near-linear savings in wall-clock time. In contrast, finetuning
benefits less dramatically since language-heavy updates dominate compute at
this stage.

Overall, these results demonstrate that Delta-LLaVA yields significant
efficiency improvements without sacrificing downstream performance, making it particularly attractive for scaling
multimodal training under limited GPU budgets.

\subsection{Delta-LLaVA with different Vision Towers Analysis}

Table \ref{224} compares the performance of different vision towers when paired with the Vicuna-7B backbone under a fixed token budget. Both CLIP-based models achieve strong results, with CLIP-336 slightly outperforming CLIP-224 on most benchmarks, particularly on GQA and POPE, while CLIP-224 shows a modest advantage on SQA. SigLIP-386, despite processing more tokens (81 vs. 64), trails behind on most tasks, especially on MME and MMB, though it remains competitive on POPE and SQA. These results highlight that increasing the resolution or token count does not guarantee better performance, and that CLIP-based towers provide a more balanced trade-off between accuracy across datasets and computational efficiency.
\begin{table}[h]
\centering
\begin{tabular}{lcccccccc}
\toprule
\textbf{Vision Tower} & \textbf{\# Token} & \textbf{GQA} & \textbf{MMB} & \textbf{MME} & \textbf{POPE} & \textbf{SQA}  & \textbf{VizWiz}  \\
\midrule
\textbf{CLIP 336} & 64 & \textbf{61.1} &  \textbf{64.94} & 1442 & \textbf{86.12}   & 67.82  & \textbf{57.2} \\
\textbf{CLIP 224} & 64 & 60.5 & 64.9 & \textbf{1448.1} &84.56 & 69.3 & 56.1  \\
\textbf{SigLIP 386} & 81 & 59.44 & 61.69 & 1345.03 &85.2 & \textbf{68.42} & 55.54 \\
\midrule
\end{tabular}
\caption{Results obtained using the Vicuna 7B and with different vision towers.}
\label{224}\vspace{-0.8em}
\end{table}
\subsection{Visual Token Pruning on Inference Efficiency}

\begin{figure*}[t]
  \centering
  \subfloat[Flickr30k CIDEr under pruning. Delta-LLaVA (solid) consistently achieves higher scores than FastV (dashed) across both 7B and 13B.]{
    \includegraphics[width=0.48\textwidth]{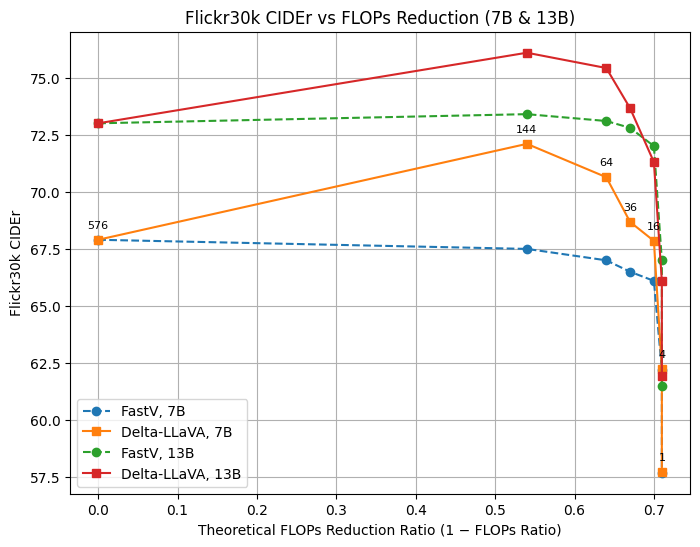}
    \label{subfig:flicker}
  }
  \hfill
  \subfloat[Nocaps CIDEr under pruning. Delta-LLaVA (solid) preserves captioning quality significantly better than FastV at both 7B and 13B.]{
    \includegraphics[width=0.48\textwidth]{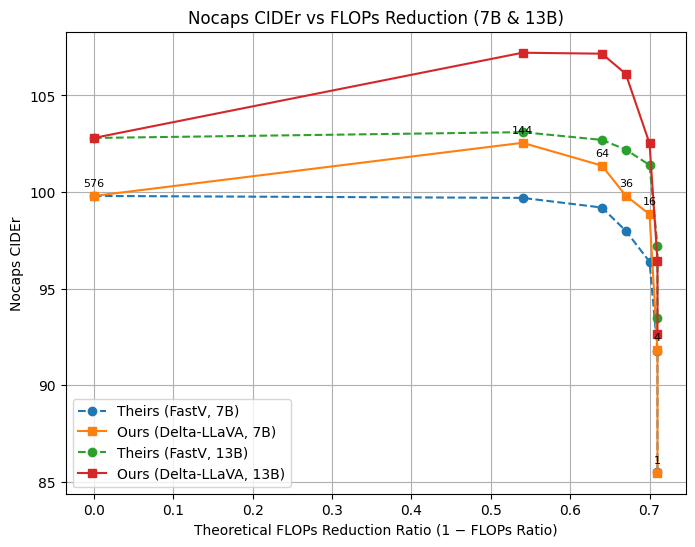}
    \label{subfig:nocap}
  }
  \caption{Comparison of captioning robustness under token pruning. Across Flickr30k and Nocaps benchmarks, Delta-LLaVA achieves a superior accuracy--efficiency trade-off relative to FastV, especially at moderate to aggressive pruning levels.}
  \label{fig:flicker-nocap}
  \vspace{-0.8em}
\end{figure*}

\smallskip
\noindent
We evaluate how pruning the number of vision tokens $K$ passed from the
vision tower to the LLM affects inference when generation is allowed to stop
naturally (averaging $\sim30$ new tokens). We sweep $K \in \{576,144,64,36,16,4,1\}$.
Figure \ref{fig:flops-gqa} reports both total compute (stacked by component) and
tokens-per-second (TPS), with TPS values annotated above each bar.

\smallskip
\noindent
\textbf{Observed behavior.}
TPS improves as $K$ decreases---from $23.96$ tok/s at $K{=}576$ to
$\approx36$--$37$ tok/s for $K \leq 16$ (peak $37.08$ tok/s at $K{=}1$;
$\sim55\%$ gain). In contrast, total FLOPs drop sharply: $7.43$ TFLOPs at
$K{=}576$, $2.84$ at $144$, and $1.74$--$1.39$ for $K \leq 36$
($\sim81\%$ reduction from $576 \to 1$). The breakdown shows that pruning
mainly reduces decoder \textit{prefill} cost (scales with $T{+}K$), falling
from $6.72$ TFLOPs ($K{=}576$) to $0.70$ TFLOPs ($K{=}1$). The
\textit{decode} stage remains nearly constant across $K$ ($\approx0.28$--$0.31$
TFLOPs for $\sim30$ generated tokens). The vision encoder is fixed by input
resolution ($\approx0.382$ TFLOPs), and the projector is negligible
($\leq0.002$ TFLOPs except $0.024$ at $K{=}576$).

\begin{table}
\centering

\small
\setlength{\tabcolsep}{5pt}
\begin{tabular}{lcccccccc}
\toprule
 & \multicolumn{4}{c|}{\textbf{LLaVA-1.5-7B}} & \multicolumn{4}{c}{\textbf{LLaVA-1.5-13B}} \\
\cmidrule(lr){2-5} \cmidrule(lr){6-9}
\textbf{\# Tokens} & \multicolumn{2}{c|}{Nocaps CIDEr} & \multicolumn{2}{c|}{Flickr30k CIDEr} & \multicolumn{2}{c|}{Nocaps CIDEr} & \multicolumn{2}{c}{Flickr30k CIDEr} \\
\midrule
 & FastV & Ours & FastV & Ours & FastV & Ours & FastV & Ours \\
\midrule
576 & 99.8 & 99.8 & 67.9 & 67.9 & 102.8 & 102.8 & 73.0 & 73.0 \\
144 & 99.7 & \textbf{102.6} & 67.5 & \textbf{72.1} & 103.1 & \textbf{107.2} & 73.4 & \textbf{76.1} \\
64  & 99.2 & \textbf{101.4} & 67.0 & \textbf{70.6} & 102.7 & \textbf{107.2} & 73.1 & \textbf{75.4} \\
36  & 98.0 & \textbf{99.8}  & 66.5 & \textbf{68.7} & 102.2 & \textbf{106.1} & 72.8 & \textbf{73.7} \\
16  & 96.4 & \textbf{98.9}  & 66.1 & \textbf{67.9} & 101.4 & \textbf{102.5} & \textbf{72.0} & 71.3 \\
4   & 91.8 & \textbf{91.9}  & 62.2 & \textbf{62.2} & \textbf{97.2}  & 96.5  & \textbf{67.0} & 66.1 \\
1   & 85.5 & \textbf{85.5}  & 57.7 & \textbf{57.7} & \textbf{93.5}  & 92.7  & 61.5 & \textbf{61.9} \\
\bottomrule

\end{tabular}
\caption{Comparison of Nocaps and Flickr30k CIDEr scores under pruning for LLaVA-1.5-7B and LLaVA-1.5-13B. We report results from FastV and our Delta-LLaVA implementation (“Ours”). Token counts $K$ correspond to the number of visual tokens retained.}\label{image_cap}\vspace{-0.8em}
\end{table}
\subsection{Performance of Delta-LLaVA in image captioning tasks}
Our results in Table \ref{image_cap} demonstrate that Delta-LLaVA outperforms FastV \cite{chen2024image} across both Nocaps and Flickr30k CIDEr benchmarks for LLaVA-1.5-7B and 13B. At baseline (576 tokens), performance is matched, but once pruning is applied our models maintain higher scores at all levels—with particularly large margins at moderate pruning (e.g., 144–64 tokens), where Delta-LLaVA achieves up to 3–5 CIDEr points higher. This indicates that our approach is especially well-suited for captioning-style tasks (Nocaps, Flickr30k), where preserving fine-grained visual grounding under compression is critical. Even under extreme pruning (K=1–4), our curves degrade more gracefully, retaining $\geq$ 85\% of baseline performance, while FastV shows sharper declines. Together, these findings highlight Delta-LLaVA as a more robust token-efficient model, delivering stronger accuracy–efficiency trade-offs for vision–language tasks where captioning fidelity and robustness to compression are essential, without sacrificing inference efficiency.

\section{Conclusion}
Our experiments demonstrate that Delta-LLaVA sets a new standard for token-efficient 
multimodal modeling. By introducing the Delta Projector, which combines low-rank 
alignment with lightweight specialization layers, the model achieves stronger reasoning 
performance and robustness under severe visual token compression. 

Across benchmarks, Delta-LLaVA sustains competitive or superior accuracy even when 
the number of visual tokens is reduced by orders of magnitude. At moderate compression 
(e.g., 144 tokens), it surpasses strong baselines on tasks such as GQA, Nocaps, and 
Flicker30K, while at aggressive settings (36–16 tokens) it attains the best or 
second-best scores on GQA, VizWiz, SQA, and POPE. Remarkably, the model remains 
resilient at extreme compression (4–1 tokens), where prior methods fail. Scaling to 
larger LLM backbones further amplifies these gains, underscoring the synergy between 
stronger language capacity and efficient visual projection. 

Efficiency analyses confirm that compressing vision tokens not only reduces FLOPs 
by up to 70\% (576 $\rightarrow$ 1 token), but also accelerates training by more than 
$4-5\times$ during pretraining and over $1.5\times$ during instruction tuning. These 
results highlight the importance of principled token formation for building scalable 
and efficient multimodal large language models. 

\bibliographystyle{plainnat} 
\bibliography{main}

@article{liang2022not,
  title={Not all patches are what you need: Expediting vision transformers via token reorganizations},
  author={Liang, Youwei and Ge, Chongjian and Tong, Zhan and Song, Yibing and Wang, Jue and Xie, Pengtao},
  journal={arXiv preprint arXiv:2202.07800},
  year={2022}
}

@inproceedings{chen2024image,
  title={An image is worth 1/2 tokens after layer 2: Plug-and-play inference acceleration for large vision-language models},
  author={Chen, Liang and Zhao, Haozhe and Liu, Tianyu and Bai, Shuai and Lin, Junyang and Zhou, Chang and Chang, Baobao},
  booktitle={European Conference on Computer Vision},
  pages={19--35},
  year={2024},
  organization={Springer}
}

@article{rao2021dynamicvit,
  title={Dynamicvit: Efficient vision transformers with dynamic token sparsification},
  author={Rao, Yongming and Zhao, Wenliang and Liu, Benlin and Lu, Jiwen and Zhou, Jie and Hsieh, Cho-Jui},
  journal={Advances in neural information processing systems},
  volume={34},
  pages={13937--13949},
  year={2021}
}

@article{liu2023visual,
  title={Visual instruction tuning},
  author={Liu, Haotian and Li, Chunyuan and Wu, Qingyang and Lee, Yong Jae},
  journal={Advances in neural information processing systems},
  volume={36},
  pages={34892--34916},
  year={2023}
}

@inproceedings{liu2024improved,
  title={Improved baselines with visual instruction tuning},
  author={Liu, Haotian and Li, Chunyuan and Li, Yuheng and Lee, Yong Jae},
  booktitle={Proceedings of the IEEE/CVF Conference on Computer Vision and Pattern Recognition},
  pages={26296--26306},
  year={2024}
}

@article{dai2020funnel,
  title={Funnel-transformer: Filtering out sequential redundancy for efficient language processing},
  author={Dai, Zihang and Lai, Guokun and Yang, Yiming and Le, Quoc},
  journal={Advances in neural information processing systems},
  volume={33},
  pages={4271--4282},
  year={2020}
}

@article{nawrot2022efficient,
  title={Efficient transformers with dynamic token pooling},
  author={Nawrot, Piotr and Chorowski, Jan and {\L}a{\'n}cucki, Adrian and Ponti, Edoardo M},
  journal={arXiv preprint arXiv:2211.09761},
  year={2022}
}

@article{chen2024llavolta,
  title={Llavolta: Efficient multi-modal models via stage-wise visual context compression},
  author={Chen, Jieneng and Ye, Luoxin and He, Ju and Wang, Zhao-Yang and Khashabi, Daniel and Yuille, Alan},
  journal={arXiv preprint arXiv:2406.20092},
  year={2024}
}

@inproceedings{cai2024matryoshka,
  title={Matryoshka multimodal models},
  author={Cai, Mu and Yang, Jianwei and Gao, Jianfeng and Lee, Yong Jae},
  booktitle={Workshop on Video-Language Models@ NeurIPS 2024},
  year={2024}
}

@article{hu2024matryoshka,
  title={Matryoshka query transformer for large vision-language models},
  author={Hu, Wenbo and Dou, Zi-Yi and Li, Liunian Harold and Kamath, Amita and Peng, Nanyun and Chang, Kai-Wei},
  journal={arXiv preprint arXiv:2405.19315},
  year={2024}
}

@inproceedings{liang2024survey,
  title={A Survey of Multimodel Large Language Models},
  author={Liang, Zijing and Xu, Yanjie and Hong, Yifan and Shang, Penghui and Wang, Qi and Fu, Qiang and Liu, Ke},
  booktitle={Proceedings of the 3rd International Conference on Computer, Artificial Intelligence and Control Engineering},
  pages={405--409},
  year={2024}
}

@article{li2023evaluating,
  title={Evaluating object hallucination in large vision-language models},
  author={Li, Yifan and Du, Yifan and Zhou, Kun and Wang, Jinpeng and Zhao, Wayne Xin and Wen, Ji-Rong},
  journal={arXiv preprint arXiv:2305.10355},
  year={2023}
}

@article{shang2024llava,
  title={Llava-prumerge: Adaptive token reduction for efficient large multimodal models},
  author={Shang, Yuzhang and Cai, Mu and Xu, Bingxin and Lee, Yong Jae and Yan, Yan},
  journal={arXiv preprint arXiv:2403.15388},
  year={2024}
}

@article{touvron2023llama,
  title={Llama: Open and efficient foundation language models},
  author={Touvron, Hugo and Lavril, Thibaut and Izacard, Gautier and Martinet, Xavier and Lachaux, Marie-Anne and Lacroix, Timoth{\'e}e and Rozi{\`e}re, Baptiste and Goyal, Naman and Hambro, Eric and Azhar, Faisal and others},
  journal={arXiv preprint arXiv:2302.13971},
  year={2023}
}

@inproceedings{liu2024mmbench,
  title={Mmbench: Is your multi-modal model an all-around player?},
  author={Liu, Yuan and Duan, Haodong and Zhang, Yuanhan and Li, Bo and Zhang, Songyang and Zhao, Wangbo and Yuan, Yike and Wang, Jiaqi and He, Conghui and Liu, Ziwei and others},
  booktitle={European conference on computer vision},
  pages={216--233},
  year={2024},
  organization={Springer}
}

@inproceedings{radford2021learning,
  title={Learning transferable visual models from natural language supervision},
  author={Radford, Alec and Kim, Jong Wook and Hallacy, Chris and Ramesh, Aditya and Goh, Gabriel and Agarwal, Sandhini and Sastry, Girish and Askell, Amanda and Mishkin, Pamela and Clark, Jack and others},
  booktitle={International conference on machine learning},
  pages={8748--8763},
  year={2021},
  organization={PmLR}
}

@article{kusupati2022matryoshka,
  title={Matryoshka representation learning},
  author={Kusupati, Aditya and Bhatt, Gantavya and Rege, Aniket and Wallingford, Matthew and Sinha, Aditya and Ramanujan, Vivek and Howard-Snyder, William and Chen, Kaifeng and Kakade, Sham and Jain, Prateek and others},
  journal={Advances in Neural Information Processing Systems},
  volume={35},
  pages={30233--30249},
  year={2022}
}

@article{li2024tokenpacker,
  title={Tokenpacker: Efficient visual projector for multimodal llm},
  author={Li, Wentong and Yuan, Yuqian and Liu, Jian and Tang, Dongqi and Wang, Song and Qin, Jie and Zhu, Jianke and Zhang, Lei},
  journal={arXiv preprint arXiv:2407.02392},
  year={2024}
}

@inproceedings{li2023blip,
  title={Blip-2: Bootstrapping language-image pre-training with frozen image encoders and large language models},
  author={Li, Junnan and Li, Dongxu and Savarese, Silvio and Hoi, Steven},
  booktitle={International conference on machine learning},
  pages={19730--19742},
  year={2023},
  organization={PMLR}
}

@article{zhang2024cls,
  title={[CLS] Attention is All You Need for Training-Free Visual Token Pruning: Make VLM Inference Faster},
  author={Zhang, Qizhe and Cheng, Aosong and Lu, Ming and Zhuo, Zhiyong and Wang, Minqi and Cao, Jiajun and Guo, Shaobo and She, Qi and Zhang, Shanghang},
  journal={arXiv preprint arXiv:2412.01818},
  year={2024}
}

@article{li2024inference,
  title={Inference optimal vlms need only one visual token but larger models},
  author={Li, Kevin Y and Goyal, Sachin and Semedo, Joao D and Kolter, J Zico},
  journal={arXiv preprint arXiv:2411.03312},
  year={2024}
}

@inproceedings{gurari2018vizwiz,
  title={Vizwiz grand challenge: Answering visual questions from blind people},
  author={Gurari, Danna and Li, Qing and Stangl, Abigale J and Guo, Anhong and Lin, Chi and Grauman, Kristen and Luo, Jiebo and Bigham, Jeffrey P},
  booktitle={Proceedings of the IEEE conference on computer vision and pattern recognition},
  pages={3608--3617},
  year={2018}
}

@inproceedings{goyal2017making,
  title={Making the v in vqa matter: Elevating the role of image understanding in visual question answering},
  author={Goyal, Yash and Khot, Tejas and Summers-Stay, Douglas and Batra, Dhruv and Parikh, Devi},
  booktitle={Proceedings of the IEEE conference on computer vision and pattern recognition},
  pages={6904--6913},
  year={2017}
}

@inproceedings{hudson2019gqa,
  title={Gqa: A new dataset for real-world visual reasoning and compositional question answering},
  author={Hudson, Drew A and Manning, Christopher D},
  booktitle={Proceedings of the IEEE/CVF conference on computer vision and pattern recognition},
  pages={6700--6709},
  year={2019}
}

@inproceedings{cha2024honeybee,
  title={Honeybee: Locality-enhanced projector for multimodal llm},
  author={Cha, Junbum and Kang, Wooyoung and Mun, Jonghwan and Roh, Byungseok},
  booktitle={Proceedings of the IEEE/CVF Conference on Computer Vision and Pattern Recognition},
  pages={13817--13827},
  year={2024}
}

@article{mikaelyan2025deltallm,
  title={DeltaLLM: Compress LLMs with Low-Rank Deltas between Shared Weights},
  author={Mikaelyan, Liana and Imani, Ayyoob and Salvaris, Mathew and Pathak, Parth and Fayyaz, Mohsen},
  journal={arXiv preprint arXiv:2501.18596},
  year={2025}
}

@article{team2024gemma,
  title={Gemma: Open models based on gemini research and technology},
  author={Team, Gemma and Mesnard, Thomas and Hardin, Cassidy and Dadashi, Robert and Bhupatiraju, Surya and Pathak, Shreya and Sifre, Laurent and Rivi{\`e}re, Morgane and Kale, Mihir Sanjay and Love, Juliette and others},
  journal={arXiv preprint arXiv:2403.08295},
  year={2024}
}

@inproceedings{zhang2025lmms,
  title={Lmms-eval: Reality check on the evaluation of large multimodal models},
  author={Zhang, Kaichen and Li, Bo and Zhang, Peiyuan and Pu, Fanyi and Cahyono, Joshua Adrian and Hu, Kairui and Liu, Shuai and Zhang, Yuanhan and Yang, Jingkang and Li, Chunyuan and others},
  booktitle={Findings of the Association for Computational Linguistics: NAACL 2025},
  pages={881--916},
  year={2025}
}

@inproceedings{agrawal2019nocaps,
  title={Nocaps: Novel object captioning at scale},
  author={Agrawal, Harsh and Desai, Karan and Wang, Yufei and Chen, Xinlei and Jain, Rishabh and Johnson, Mark and Batra, Dhruv and Parikh, Devi and Lee, Stefan and Anderson, Peter},
  booktitle={Proceedings of the IEEE/CVF international conference on computer vision},
  pages={8948--8957},
  year={2019}
}

@inproceedings{plummer2015flickr30k,
  title={Flickr30k entities: Collecting region-to-phrase correspondences for richer image-to-sentence models},
  author={Plummer, Bryan A and Wang, Liwei and Cervantes, Chris M and Caicedo, Juan C and Hockenmaier, Julia and Lazebnik, Svetlana},
  booktitle={Proceedings of the IEEE international conference on computer vision},
  pages={2641--2649},
  year={2015}
}

@inproceedings{li2024enhancing,
  title={Enhancing Visual Information Extraction with Large Language Models Through Layout-Aware Instruction Tuning},
  author={Li, Teng and Wang, Jiapeng and Jin, Lianwen},
  booktitle={Chinese Conference on Pattern Recognition and Computer Vision (PRCV)},
  pages={276--289},
  year={2024},
  organization={Springer}
}

@article{alayrac2022flamingo,
  title={Flamingo: a visual language model for few-shot learning},
  author={Alayrac, Jean-Baptiste and Donahue, Jeff and Luc, Pauline and Miech, Antoine and Barr, Iain and Hasson, Yana and Lenc, Karel and Mensch, Arthur and Millican, Katherine and Reynolds, Malcolm and others},
  journal={Advances in neural information processing systems},
  volume={35},
  pages={23716--23736},
  year={2022}
}

@article{sun2023eva,
  title={Eva-clip: Improved training techniques for clip at scale},
  author={Sun, Quan and Fang, Yuxin and Wu, Ledell and Wang, Xinlong and Cao, Yue},
  journal={arXiv preprint arXiv:2303.15389},
  year={2023}
}

@article{tschannen2025siglip,
  title={Siglip 2: Multilingual vision-language encoders with improved semantic understanding, localization, and dense features},
  author={Tschannen, Michael and Gritsenko, Alexey and Wang, Xiao and Naeem, Muhammad Ferjad and Alabdulmohsin, Ibrahim and Parthasarathy, Nikhil and Evans, Talfan and Beyer, Lucas and Xia, Ye and Mustafa, Basil and others},
  journal={arXiv preprint arXiv:2502.14786},
  year={2025}
}

\end{document}